\documentclass{article}

\usepackage[final]{nips_2016}

\usepackage[utf8]{inputenc} 
\usepackage[T1]{fontenc}    
\usepackage{bm}             
\usepackage{hyperref}       
\usepackage{url}            
\usepackage{booktabs}       
\usepackage{amsfonts}       
\usepackage{amsmath}        
\usepackage{amssymb}        
\usepackage{nicefrac}       
\usepackage{microtype}      
\usepackage{todonotes}
\usepackage{setspace}
\usepackage{blindtext}

\newcommand{\Reals}{\mathbb{R}}

\newcommand{\T}{\top}
\newcommand{\tra}[1]{#1^{\T}}

\newcommand{\I}{{-1}}
\newcommand{\inv}[1]{#1^{\I}}

\newcommand{\vect}[1]{\bm{\mathrm{#1}}}
\newcommand{\matr}[1]{\mathrm{#1}}

\newcommand{\tvect}[1]{\tra{\vect{#1}}}
\newcommand{\tmatr}[1]{\tra{\matr{#1}}}
\newcommand{\imatr}[1]{\inv{\matr{#1}}}

\newcommand{\eye}[1]{\matr{I}_{#1}}
\newcommand{\onesv}[1]{\vect{1}_{#1}}

\newcommand{\Tr}[1]{\operatorname{Tr}[#1]}

\newcommand{\dotprod}[2]{\langle #1 , #2 \rangle}
\newcommand{\kronprod}[2]{( #1 \otimes #2 )}

\newcommand{\Normal}{\mathcal{N}}

\newcommand{\given}{\mid}
\newcommand{\sampled}{\sim}

\newcommand{\Cov}{\operatorname{Cov}}

\newcommand{\Exp}{\mathbb{E}}
\newcommand{\KL}{\operatorname{KL}}

\newcommand{\Indicator}[1]{\mathbb{I}[ #1 ]}

\newcommand{\tabresult}[2]{#1 ($\pm$ #2)}

\newcommand{\expk}{\Exp_{q(\vect{x}_i)}[\vect{k}_i]}
\newcommand{\expkt}{\Exp_{q(\vect{x}_i)}[\tvect{k}_i]}
\newcommand{\expkkt}{\Exp_{q(\vect{x}_i)}[\vect{k}_i \tvect{x}_i]}

\newcommand{\Note}[2]{} 
\newcommand{\SideNote}[2]{} 
\renewcommand{\Note}[2]{\todo[color=#1,size=\small, inline=true]{#2}} 
\renewcommand{\SideNote}[2]{\todo[color=#1,size=\small]{#2}} 

\let\oldbibliography\thebibliography
\renewcommand{\thebibliography}[1]{\oldbibliography{#1}
\setlength{\itemsep}{0.3pt}}

\title{Disease Trajectory Maps}

\author{
  Peter Schulam \\
  Dept. of Computer Science \\
  Johns Hopkins University \\
  Baltimore, MD 21218 \\
  \texttt{pschulam@cs.jhu.edu} \\
  \And
  Raman Arora \\
  Dept. of Computer Science \\
  Johns Hopkins University \\
  Baltimore, MD 21218 \\
  \texttt{arora@cs.jhu.edu} \\
}

\begin{document}

\maketitle

\vspace*{-15pt}
\begin{abstract}
  Medical researchers are coming to appreciate that many diseases are
  in fact complex, heterogeneous syndromes composed of subpopulations
  that express different variants of a related complication. Time
  series data extracted from individual electronic health records
  (EHR) offer an exciting new way to study subtle differences in the
  way these diseases progress over time. In this paper, we focus on
  answering two questions that can be asked using these databases of
  time series. First, we want to understand whether there are
  individuals with similar \emph{disease trajectories} and whether
  there are a small number of \emph{degrees of freedom} that account
  for differences in trajectories across the population. Second, we
  want to understand how important clinical outcomes are associated
  with disease trajectories. To answer these questions, we propose the
  Disease Trajectory Map (DTM), a novel probabilistic model that
  learns low-dimensional representations of sparse and irregularly
  sampled time series. We propose a stochastic variational inference
  algorithm for learning the DTM that allows the model to scale to
  large modern medical datasets. To demonstrate the DTM, we analyze
  data collected on patients with the complex autoimmune disease,
  scleroderma. We find that DTM learns meaningful representations of
  disease trajectories that the representations are significantly
  associated with important clinical outcomes.
\end{abstract}

\vspace*{-5pt}
\vspace*{-8pt}
\section{Introduction}
\label{sec:introduction}
\vspace*{-5pt}

Time series data is becoming increasingly important in medical
research and practice. This is due, in part, to the growing adoption
of electronic health records (EHRs), which capture snapshots of an
individual's state over time. These snapshots include clinical
observations (apparent symptoms and vital sign measurements),
laboratory test results, and treatment information. In parallel,
medical researchers are beginning to recognize and appreciate that
many diseases are in fact complex, highly heterogeneous syndromes
\citep{craig2008complex} and that individuals may belong to disease
subpopulations or \emph{subtypes} that express similar sets of
symptoms over time (see e.g. \citet{saria2015subtyping}). Examples of
such diseases include asthma \citep{lotvall2011asthma}, autism
\citep{wiggins2012support}, and COPD \citep{castaldi2014cluster}. The
data captured in EHRs can help better understand these complex
diseases. EHRs contain a multitude of types of observations and the
ability to track their progression can help bring in to focus the
subtle differences across individual disease expression.

In this paper, we focus on two exploratory questions that we can begin
to answer using repositories of biomedical time series data. First, we
want to discover whether there are individuals with similar
\emph{disease trajectories} and whether there are a small number of
\emph{degrees of freedom} that account for differences across a
heterogeneous population. A better understanding of the types of
trajectories and how they differ can yield insights into the
biological underpinnings of the disease. In turn, this may motivate
new targeted therapies. In the clinic, physicians can analyze an
individual's clinical history to better understand the ``flavor'' of
the disease being expressed and can use this knowledge to make more
accurate prognoses and guide treatment decisions. Second, we would
like to know whether individuals with similar clinical outcomes
(e.g. death, severe organ damage, or development of comorbidities)
have similar disease trajectories. In complex diseases, individuals
are often at risk of developing a number of severe complications and
clinicians rarely have access to accurate prognostic
biomarkers. Discovering associations between target outcomes and
trajectory patterns may both generate new hypotheses regarding the
causes of these outcomes and help clinicians to better anticipate the
event using an individual's clinical history.

\vspace*{-7pt}
\paragraph{Contributions.}
Our approach to simultaneously answering these questions is to embed
individual disease trajectories into a low-dimensional vector space
wherein similarity in the embedded space implies that two individuals
have similar trajectories. Such an embedding would naturally answer
our first question, and the results could also be used to answer the
second by comparing distributions over embeddings across groups
defined by different outcomes. To produce such an embedding, we
introduce a novel probabilistic model of biomedical time series data,
which we term the Disease Trajectory Map (DTM). In particular, the DTM
models the trajectory of a single \emph{clinical marker}, which is an
observation or measurement recorded over time by clinicians that are
used to track the progression of a disease (see
e.g. \citet{schulam2015clustering}). Examples of clinical markers are
pulmonary function tests or creatinine laboratory test results, which
track lung and kidney function respectively. The DTM discovers
low-dimensional (2D or 3D) latent representations of clinical marker
trajectories that are easy to visualize. Moreover, the model learns an
expressive family of distributions over trajectories that is
parameterized by the low-dimensional representation. This allows the
DTM to capture a wide variety of trajectory shapes, making it suitable
for studying complex diseases where expression varies widely across
the population. We describe a stochastic variational inference
algorithm for estimating the posterior distribution over the
parameters and individual-specific representations, which allows our
model to be easily applied to large biomedical datasets. To
demonstrate the DTM, we analyze clinical marker data collected on
individuals with the complex autoimmune disease scleroderma (see
e.g. \citet{allanore2015systemic}). We find that the learned
representations capture interesting subpopulations consistent with
previous findings, and that the representations suggest associations
with important clinical outcomes.

\vspace*{-6pt}
\subsection{Background and Related Work}
\label{sec:background-and-related-work}
\vspace*{-5pt}

Clinical marker data extracted from EHRs is a by-product of an
individual's interactions with the healthcare system. As a result, the
time series are often irregularly sampled (the time between samples
varies within and across individuals), and may be extremely sparse (it
is not unusual to have a single observation for an individual). To aid
the following discussion, we briefly introduce notation for this type
of data. We use $m$ to denote the number of individual disease
trajectories recorded in a given dataset. For each individual, we use
$n_i$ to denote the number of observations. We collect the observation
times for subject $i$ into a column vector $\vect{t}_i$ (sorted in
non-decreasing order) and the corresponding measurements into a column
vector $\vect{y}_i$:
$\vect{t}_i \triangleq \tra{[t_{i 1}, \ldots, t_{i n_i}]}$ and
$\vect{y}_i \triangleq \tra{[y_{i 1}, \ldots, y_{i n_i}]}$.
Our goal is to embed the pair $(\vect{t}_i, \vect{y}_i)$ into a
low-dimensional vector space wherein similarity between two embeddings
$(\vect{x}_i, \vect{x}_j)$ implies that the trajectories have similar
shapes. This is commonly done using \emph{basis representations} of
the trajectories.

\vspace*{-10pt}
\paragraph{Fixed basis representations.}
In the statistics literature, trajectory data is often referred to as
\emph{unbalanced longitudinal data}, and it is commonly analyzed in
that community using linear mixed models (LMMs)
\citep{verbeke2009linear}. In their simplest form, LMMs assume the
following probabilistic model:
\vspace{-4pt}
\begin{align}
  \vect{w}_i \given \matr{\Sigma} &\sampled \Normal(\matr{0}, \matr{\Sigma})
  \,\,\, , \,\,\,
  \vect{y}_i \given \matr{B}_i, \vect{w}_i, \mu, \sigma^2
  \sampled \Normal(\mu + \matr{B}_i \vect{w}_i, \sigma^2 \eye{n_i}).
\end{align}
The matrix $\matr{B}_i \in \Reals^{n_i \times d}$ is known as the
\emph{design matrix}, and can be used to capture non-linear
relationships between the observation times $\vect{t}_i$ and
measurements $\vect{y}_i$. Its rows are comprised of $d$-dimensional
basis expansions of each observation time
$\matr{B}_i = \tra{[ \vect{b}(t_{i1}), \cdots, \vect{b}(t_{i n_i})
  ]}$. Common choices of $\vect{b}(\cdot)$ include polynomials,
splines, wavelets, and Fourier series. The particular basis used is
often carefully crafted by the analyst depending on the nature of the
trajectories and on the desired structure (e.g. invariance to
translations and scaling) in the
representation~\citep{brillinger2001time}. The design matrix can
therefore make the LMM remarkably flexible despite its simple
parametric probabilistic assumptions. Moreover, the prior over
$\vect{w}_i$ and the conjugate likelihood make it straightforward to
fit $\mu$, $\matr{\Sigma}$, and $\sigma^2$ using EM or Bayesian
posterior inference.

After estimating the model parameters, we can estimate the
coefficients $\vect{w}_i$ of a given clinical marker trajectory using
the posterior distribution, which embeds the trajectory in a Euclidean
space. To flexibly capture complex trajectory shapes, however, the
basis must be high-dimensional, which makes interpretability of the
embeddings challenging. We can use low-dimensional summaries such as
the projection on to a principal subspace, but these are not
necessarily substantively meaningful. Indeed, much research has gone
into developing principal direction post-processing techniques
(e.g. \citet{kaiser1958varimax}) or alternative estimators that
enhance interpretability (e.g. \cite{carvalho2012high}).

\vspace*{-10pt}
\paragraph{Data-adaptive basis representations.}
A set of related, but more flexible, techniques comes from functional
data analysis where observations are functions (i.e. trajectories)
assumed to be sampled from a stochastic process and the goal is to
find a parsimonious representation for the
data~\citep{ramsay2002applied}.  Functional principal component
analysis (FPCA), one of the most standard techniques in functional
data analysis, expresses functional data in the orthonormal basis
given by the eigenfunctions of the auto-covariance operator. This
representation is optimal in the sense that no other representation
captures more variation~\citep{ramsay2006functional}. The idea itself
can be traced back to early independent work by Karhunen and Loeve and
is also referred to as the Karhunen-Loeve
expansion~\citep{watanabe1965karhunen}. While numerous variants of
FPCA have been proposed, the one that is most relevant to the problem
at hand is that of sparse FPCA~\citep{castro1986principal,
  rice2001nonparametric} where we allow sparse irregularly sampled
data as in longitudinal data analysis. To deal with the sparsity,
\citet{rice2001nonparametric} proposed the mixed effect model which
leverages statistical strength from all observations for function
estimation. The mixed effect model often suffers from numerical
instability of covariance matrices in high dimensions;
\citet{james2000principal} addressed this by constraining the rank of
the covariance matrices---this is often referred to as the reduced
rank model. The reduced rank model was further extended
by~\citet{zhou2008joint} to a two-dimensional sparse principal
component model. Although the reduced rank model embeds trajectories
using a data-driven basis, the basis is restricted to lie in a linear
subspace of a fixed basis, which may be overly restrictive. Other
approaches to learning a functional basis include Bayesian estimation
of B-spline parameters (e.g. \citep{bigelow2012bayesian}) and placing
priors over reproducing kernel Hilbert spaces
(e.g. \citep{maclehose2009nonparametric}). Although flexible, these
two approaches do not learn a compact representation.

\vspace*{-10pt}
\paragraph{Cluster-based representations.}
Mixture models and clustering approaches are also commonly used to
represent and discover structure in time series
data. \citet{marlin2012unsupervised} cluster time series data from the
ICU using a mixture model and use cluster membership to predict
outcomes. \citet{schulam2015framework} describe a probabilistic model
that represents trajectories using a hierarchy of features, which
includes ``subtype'' or cluster membership. LMMs have also been
extended to have nonparametric Dirichlet process priors over the
coefficients (e.g. \citet{kleinman1998semiparametric}), which
implicitly induce clusters in the data. Although these approaches
flexibly model trajectory data, the structure they recover is a
partition, which does not allow us to compare all trajectories in a
coherent way as we can in a vector space.

\vspace*{-10pt}
\paragraph{Lexicon-based representations.}
Another line of research has investigated the discovery of motifs or
repeated patterns in continuous time-series data for the purposes of
succinctly representing the data as a string of elements of the
discovered lexicon. These include efforts in the speech processing
community to identify sub-word units (parts of the words at the same
level as phonemes) in a data-driven
manner~\citep{varadarajan2008unsupervised,levin2013fixed}.
In computational healthcare, \citet{saria2011discovering} propose a
method for discovering deformable motifs that are repeated in
continuous time-series data. These methods are, in spirit, similar to
discretization approaches such as symbolic aggregate approximation
(SAX)~\citep{lin2007experiencing} and piecewise aggregate
approximation (PAX)~\citep{keogh2001locally} that are popular in data
mining, and aim to find compact description of sequential data,
primarily for the purposes of indexing, search, anomaly detection, and
information retrieval. The focus in this paper is to learn
representations for entire trajectories rather than discover a
lexicon. Furthermore, we are interested in learning a representation
for individuals and capturing latent similarities between two patients
in terms of Euclidean distances in the learnt representation.

\vspace*{-5pt}
\vspace*{-5pt}
\section{Disease Trajectory Maps}
\label{sec:the-disease-trajectory-map}
\vspace*{-5pt}

To motivate Disease Trajectory Maps, we begin from the reduced-rank
formulation of linear mixed models as proposed by
\citet{james2000principal}. In particular, let $\mu \in \Reals$ be the
marginal mean of the observations, $\matr{F} \in \Reals^{d \times q}$
be a rank-$q$ matrix, and $\sigma^2$ be the variance of measurement
errors. As a reminder, $\vect{y}_i \in \Reals^{n_i}$ denotes the
vector of observed trajectory measurements,
$\matr{B}_i \in \Reals^{n_i \times d}$ denotes the subject's design
matrix, and $\vect{x}_i \in \Reals^q$ denotes the representation or
\emph{embedding} of the subject. We begin with the reduced-rank
conditional model:
$\vect{y}_i \given \matr{B}_i, \vect{x}_i, \mu, \matr{F}, \sigma^2 \sampled \Normal(\mu + \matr{B}_i \matr{F} \vect{x}_i, \sigma^2 \eye{n_i}).$
In the reduced-rank model, we assume an isotropic normal prior over
$\vect{x}_i$ and marginalize to obtain the observed-data
log-likelihood, which is then optimized with respect to
$\matr{F}$. Here, just as was done by \citet{lawrence2004gaussian} to
derive the GPLVM, we swap the marginalization for optimization and
vice versa. By assuming a normal prior
$\Normal(\vect{0}, \alpha \eye{k})$ over the rows of $\matr{F}$ and
marginalizing we obtain:
\vspace{-2pt}
\begin{align}
  \vect{y}_i \given \matr{B}_i, \vect{x}_i, \mu, \sigma^2, \alpha
  \sampled \Normal(\mu, \alpha \dotprod{\vect{x}_i}{\vect{x}_i} \matr{B}_i \tmatr{B}_i + \sigma^2 \eye{n_i}).
\end{align}
Note that by marginalizing over $\matr{F}$, we induce a joint
distribution over all trajectories in the dataset. Moreover, this
joint distribution is a Gaussian process with mean $\mu$ and the
following covariance function defined across trajectories:
\vspace{-2pt}
\begin{align}
  \Cov(\vect{y}_i, \vect{y}_j \given \matr{B}_i, \matr{B}_j, \vect{x}_i, \vect{x}_j, \mu, \sigma^2, \alpha)
  = \alpha \dotprod{\vect{x}_i}{\vect{x}_j} \matr{B}_i \tmatr{B}_j + \Indicator{i = j}\, (\sigma^2 \eye{n_i}).
\end{align}
This reformulation of the reduced-rank LMM suggests a natural
alternative to learning the representations $\vect{x}_i$ of the
subjects. Just as in the GPLVM, we can maximize the log-probability of
all trajectories with respect to the hyperparameters
$\{ \mu, \sigma^2, \alpha \}$ and the representations
$\{\vect{x}_i : i \in [m]\}$. More importantly, however, the
reformulation allows us to relate the representation to the basis
coefficients non-linearly by using the ``kernel trick'' to
reparameterize the covariance function. Let $k(\cdot, \cdot)$ denote a
non-linear kernel defined over the representations with parameters
$\vect{\theta}$, then we have:
\vspace{-2pt}
\begin{align}
  \Cov(\vect{y}_i, \vect{y}_j \given \matr{B}_i, \matr{B}_j, \vect{x}_i, \vect{x}_j, \mu, \sigma^2, \vect{\theta})
  = k(\vect{x}_i, \vect{x}_j) \matr{B}_i \tmatr{B}_j + \Indicator{i = j}\, (\sigma^2 \eye{n_i}).
\end{align}
Let
$\vect{y} \triangleq \tra{[\tra{\vect{y}_1}, \ldots,
  \tra{\vect{y}_m}]}$ denote the column vector obtained by
concatenating the measurement vectors from each trajectory. The joint
distribution over $\vect{y}$ is a multivariate normal:
\vspace{-2pt}
\begin{align}
  \vect{y} \given \matr{B}_{1:m}, \vect{x}_{1:m}, \mu, \sigma^2, \vect{\theta}
  \sampled
  \Normal( \mu, \matr{\Sigma}_{\text{DTM}} + \sigma^2 \eye{n} ),
\end{align}
where $\matr{\Sigma}_{\text{DTM}}$ is a full-rank covariance matrix
that depends on the design matrices $\matr{B}_{1:m}$, the
representations $\vect{x}_{1:m}$, and the kernel $k(\cdot, \cdot)$. In
particular, $\matr{\Sigma}_{\text{DTM}}$ is a block-structured matrix
with $m$ row blocks and $m$ column blocks. The block at the
$i^{\text{th}}$ row and $j^{\text{th}}$ column is the covariance
between $\vect{y}_i$ and $\vect{y}_j$ defined above. We complete the
model by placing isotropic Gaussian priors over $\vect{x}_i$. Note
that this model is similar to the Bayesian GPLVM
\citep{titsias2010bayesian}, but models functional data instead of
finite-dimensional vectors.

\vspace*{-5pt}
\subsection{Learning and Inference in the DTM}
\label{sec:learning-the-dtm}
\vspace*{-5pt}

As formulated, the model will scale poorly to large
datasets. Inference within each iteration of an optimization
algorithm, for example, requires storing and inverting
$\matr{\Sigma}_{\text{DTM}}$, which requires $O(n^2)$ space and
$O(n^3)$ time respectively, where $n \triangleq \sum_{i=1}^m n_i$ is
the number of clinical marker observations. For modern datasets, where
$n$ can be in the hundreds of thousands or millions, this is
unacceptable. In this section, we approximate the log-likelihood using
techniques from \citet{hensman2013gaussian} that allows us to apply
stochastic variational inference (SVI)~\citep{hoffman2013stochastic}.

Recent work in scaling Gaussian processes to large datasets focuses on
the idea of \emph{inducing
  points}~\citep{snelson2005sparse,titsias2009variational}, which are
a relatively small number of artificial observations of the Gaussian
process that act as a bottleneck and approximately capture the
information contained in the training data. Let
$\vect{f} \in \Reals^m$ denote observations of the GP at inputs
$\{\vect{x}_i\}_{i=1}^m$ and $\vect{u} \in \Reals^p$ denote inducing
points at inputs
$\{\vect{z}_i\}_{i=1}^p$. \citet{titsias2009variational} constructs
the inducing points as variational parameters by introducing an
augmented probability model:
\vspace{-2pt}
\begin{align}
  \vect{u} \sampled \Normal(\vect{0}, \matr{K}_{pp})
  \,\,\, , \,\,\,
  \vect{f} \given \vect{u} \sampled
  \Normal(\matr{K}_{mp} \imatr{K}_{pp} \vect{u}, \tilde{\matr{K}}_{mm}),
\end{align}
where $\matr{K}_{pp}$ is the Gram matrix between inducing points,
$\matr{K}_{mm}$ is the Gram matrix between observations,
$\matr{K}_{mp}$ is the cross Gram matrix between observations and
inducing points, and
$\tilde{\matr{K}}_{mm} \triangleq \matr{K}_{mm} - \matr{K}_{mp}
\inv{\matr{K}}_{pp} \matr{K}_{pm}$. \citet{titsias2009variational}
then marginalizes over $\vect{u}$ to construct a low-rank approximate
covariance matrix, which is computationally cheaper to invert using
the Woodbury identity. \citet{hensman2013gaussian} extends these ideas
by maintaining a variational distribution over $\vect{u}$ that
d-separates the observations and satisfies the conditions required to
apply SVI~\citep{hoffman2013stochastic}. Let
$\vect{y}_f = \vect{f} + \vect{\epsilon}$ where $\vect{\epsilon}$ is
iid Gaussian noise with variance $\sigma^2$, then the key result from
\citet{hensman2013gaussian} that we use here is the following bound:
\vspace*{-2pt}
\begin{align}
  \label{eq:hensman-bound}
  \textstyle
  \log p( \vect{y}_f \given \vect{u} )
  \ge \sum_{i=1}^m \Exp_{p(f_i \given \vect{u})} [ \log p( y_{fi} \given f_i) ].
\end{align}
In the interest of space, we refer the interested reader to
\citet{hensman2013gaussian} for details.


\vspace*{-5pt}
\paragraph{DTM evidence lower bound.}
When marginalizing over the rows of $\matr{F}$, we induced a Gaussian
process over the trajectories, but by doing so we implicitly induced a
Gaussian process over the subject-specific basis coefficients. Let
$\vect{w}_i \triangleq \matr{F} \vect{x}_i \in \Reals^d$ denote the
curve weights implied by the mapping $\matr{F}$ and representation
$\vect{x}_i$, and let $\vect{w}_{:,k}$ for $k \in [d]$ denote the
$k^\text{th}$ coefficient of all subjects in the dataset. After
marginalizing the $k^\text{th}$ row of $\matr{F}$ and applying the
kernel trick, we see that the vector of coefficients $\vect{w}_{:,k}$
has a Gaussian process distribution with mean $0$ and covariance
function: $\Cov(w_{ik}, w_{jk}) = \alpha k(\vect{x}_i,
\vect{x}_j)$. Moreover, the Gaussian processes across coefficients are
statistically independent of one another. To lower bound the DTM
log-likelihood, we introduce $p$ inducing points $\vect{u}_k$ for each
vector of coefficients $\vect{w}_{:,k}$ with shared inducing point
inputs $\{\vect{z}_i\}_{i=1}^p$. To refer to all inducing points
simultaneously, we will use
$\matr{U} \triangleq [ \vect{u}_1, \ldots, \vect{u}_d ]$ and
$\vect{u}$ to denote the ``vectorized'' form of $\matr{U}$ obtained by
stacking its columns. Applying the bound in (\ref{eq:hensman-bound})
we have: \vspace{-2pt}
\begin{align}
  \nonumber
  \log & p(\vect{y} \given \vect{u}, \vect{x}_{1:m})
  \ge \sum_{i=1}^m \Exp_{p(\vect{w}_i \given \vect{u}, \vect{x}_i)}[ \log p(\vect{y}_i \given \vect{w}_i) ] \\
  \label{eq:conditional-logl-lower-bound}
  &= \sum_{i=1}^m \log \Normal(\vect{y}_i \given \mu + \matr{B}_i \tmatr{U} \imatr{K}_{pp} \vect{k}_i, \sigma^2 \eye{n_i})
                 - \frac{\tilde{k}_{ii}}{2\sigma^2} \Tr{\tmatr{B}_i \matr{B}_i}
  \triangleq \sum_{i=1}^m \log \tilde{p}(\vect{y}_i \given \vect{u}, \vect{x}_i),
\end{align}
where
$\vect{k}_i \triangleq [k(\vect{x}_i, \vect{z}_1), \ldots,
k(\vect{x}_i, \vect{z}_p)]^\T$ and $\tilde{k}_{ii}$ is the
$i^\text{th}$ diagonal element of $\tilde{\matr{K}}_{mm}$. We can then
construct the variational lower bound on $\log p(\vect{y})$:
\vspace{-2pt}
\begin{align}
  \log p(\vect{y})
  &\ge \Exp_{q(\vect{u}, \vect{x}_{1:m})}[ \log p(\vect{y} \given \vect{u}, \vect{x}_{1:m}) ] - \KL( q(\vect{u}, \vect{x}_{1:m}) \, \| \,\, p(\vect{u}, \vect{x}_{1:m}) ) \\
  \textstyle
  &\ge \sum_{i=1}^m \Exp_{q(\vect{u}, \vect{x}_i)}[ \log \tilde{p}(\vect{y}_i \given \vect{u}, \vect{x}_i) ] - \KL( q(\vect{u}, \vect{x}_{1:m}) \, \| \,\, p(\vect{u}, \vect{x}_{1:m}) ),
\end{align}
where we use the lower bound in
(\ref{eq:conditional-logl-lower-bound}). Finally, to make the lower
bound concrete we specify the variational distribution
$q(\vect{u}, \vect{x}_{1:m})$ to be a product of independent
multivariate normal distributions:
\vspace{-2pt}
\begin{align}
  \textstyle
  q(\vect{u}, \vect{x}_{1:M})
  \triangleq \Normal(\vect{u} \given \vect{m}, \matr{S})
  \prod_{i=1}^m \Normal(\vect{x}_i \given \vect{m}_i, \matr{S}_i),
\end{align}
where the variational parameters to be fit are $\vect{m}$, $\matr{S}$,
and $\{\vect{m}_i, \matr{S}_i\}_{i=1}^m$.

\paragraph{Stochastic optimization of the lower bound.}
To apply SVI, we must be able to compute the gradient of the expected
value of $\log \tilde{p}(\vect{y}_i \given \vect{u}, \vect{x}_i)$
under the variational distributions. Because $\vect{u}$ and
$\vect{x}_i$ are assumed to be independent in the variational
posteriors, we can analyze the expectation in either order. Fix
$\vect{x}_i$, then we see that
$\log \tilde{p}(\vect{y}_i \given \vect{u}, \vect{x}_i)$ depends on
$\vect{u}$ only through the mean of the Gaussian density, which is a
quadratic term in log likelihood. Because $q(\vect{u})$ is
multivariate normal, we can compute the expectation in closed form.
\vspace{-2pt}
\begin{align*}
  \Exp_{q(\vect{u})} [ \log \tilde{p}(\vect{y}_i \given \vect{u}, \vect{x}_i) ]
  &= \Exp_{q(\matr{U})} [ \log \Normal(\vect{y}_i \given \mu + \kronprod{\matr{B}_i}{\tvect{k}_i \imatr{K}_{pp}} \vect{u}, \sigma^2 \eye{n_i}) ]
     - \frac{\tilde{k}_{ii}}{2 \sigma^2} \Tr{\tmatr{B}_i \matr{B}_i} \\
  &= \log \Normal(\vect{y}_i \given \mu + \matr{C}_i \vect{m}, \sigma^2 \eye{n_i}) ]
     - \frac{1}{2 \sigma^2} \Tr{\matr{S} \tmatr{C}_i \matr{C}_i}
     - \frac{\tilde{k}_{ii}}{2 \sigma^2} \Tr{\tmatr{B}_i \matr{B}_i},
\end{align*}
where we have defined
$\matr{C}_i \triangleq \kronprod{\matr{B}_i}{\tvect{k}_i
  \imatr{K}_{pp}}$ to be the \emph{extended design matrix} and
$\otimes$ is the Kronecker product. We now need to compute the
expectation of this expression with respect to $q(\vect{x}_i)$, which
entails computing the expectations of $\vect{k}_i$ (a vector) and
$\vect{k}_i \tvect{k}_i$ (a matrix). In this paper, we assume an RBF
kernel, and so the elements of the vector and matrix are all
exponentiated quadratic functions of $\vect{x}_i$. This makes the
expectations straightforward to compute given that $q(\vect{x}_i)$ is
multivariate normal.\footnote{Other kernels can be used instead, but
  the expectations may not have closed form expressions.} We therefore
see that the expected value of $\log \tilde{p}(\vect{y}_i)$ can be
computed in closed form under the assumed variational distribution.

We use the standard SVI algorithm to optimize the lower bound. We
subsample the data, optimize the likelihood of each example in the
batch with respect to the variational parameters over the
representation ($\vect{m}_i$, $\matr{S}_i$), and compute approximate
gradients of the global variational parameters ($\vect{m}$,
$\matr{S}$) and the hyperparameters. The likelihood term is conjugate
to the prior over $\matr{u}$, and so we can compute the natural
gradients with respect to the global variational parameters $\vect{m}$
and $\matr{S}$
\citep{hoffman2013stochastic,hensman2013gaussian}. Additional details
on the approximate objective and the gradients required for SVI are
given in the supplement. We provide details on initialization,
minibatch selection, and learning rates for our experiments in Section
\ref{sec:experiments}.

\vspace*{-5pt}
\paragraph{Inference on new trajectories.}
The variational distribution over the inducing point values $\vect{u}$
can be used to approximate a \emph{posterior process} over the basis
coefficients $\vect{w}_i$ \citep{hensman2013gaussian}. Therefore,
given a representation $\vect{x}_i$, we have that
\vspace{-3pt}
\begin{align}
  w_{ik} \given \vect{x}_i, \vect{m}, \vect{S}
  \sampled \Normal( \tvect{k}_i \imatr{K}_{pp} \vect{m}_{k}, \tilde{k}_{ii} + \tvect{k}_i \imatr{K}_{pp} \matr{S}_{kk} \imatr{K}_{pp} \vect{k}_i ),
\end{align}
where $\vect{m}_{k}$ is the approximate posterior mean of the
$k^\text{th}$ column of $\matr{U}$ and $\matr{S}_{kk}$ is its
covariance. The approximate joint posterior distribution over all
coefficients can be shown to be multivariate normal. Let
$\vect{\mu}(\vect{x}_i)$ be the mean of this distribution given
representation $\vect{x}_i$ and $\matr{\Sigma}(\vect{x}_i)$ be the
covariance, then the posterior predictive distribution over a new
trajectory $\vect{y}_*$ given the representation $\vect{x}_*$ is
\vspace{-3pt}
\begin{align}
  \vect{y}_* \given \vect{x}_*
  \sampled \Normal( \mu + \matr{B}_* \vect{\mu}(\vect{x}_*), \matr{B}_* \matr{\Sigma}(\vect{x}_*) \tmatr{B}_* + \sigma^2 \eye{n_*}.
\end{align}
We can then approximately marginalize with respect to the prior over
$\vect{x}_*$ or a variational approximation of the posterior given a
partial trajectory using a Monte Carlo estimate.

\vspace*{-5pt}
\vspace*{-5pt}
\section{Experiments}
\label{sec:experiments}
\vspace*{-7pt} 

We now use DTM to analyze clinical marker trajectories
of individuals with the autoimmune disease, scleroderma
\citep{allanore2015systemic}. Scleroderma is a heterogeneous and
complex chronic autoimmune disease. It can potentially affect many of
the visceral organs, such as the heart, lungs, kidneys, and
vasculature. Any given individual may experience only a subset of
complications, and the timing of the symptoms relative to disease
onset can vary considerably across individuals. Moreover, there are no
known biomarkers that accurately predict an individual's disease
course. Clinicians and medical researchers are therefore interested in
characterizing and understanding disease progression
patterns. Moreover, there are a number of clinical outcomes
responsible for the majority of morbidity among patients with
scleroderma. These include congestive heart failure, pulmonary
hypertension and pulmonary arterial hypertension, gastrointestinal
complications, and myositis \citep{varga2012scleroderma}. We use the
DTM to study associations between these outcomes and disease
trajectories.

We study two scleroderma clinical markers. The first is the percent of
predicted forced vital capacity (PFVC): a pulmonary function test
result measuring lung function. PFVC is recorded as percentage points,
and a higher value (near 100) indicates that the individual's lungs
are functioning as expected. The second clinical marker that we study
is the total modified Rodnan skin score (TSS). Scleroderma is named
after its effect on the skin, which becomes hard and fibrous during
periods of high disease activity. Because it is the most clinically
apparent symptom, many of the current sub-categorizations of
scleroderma depend on an individual's pattern of skin disease activity
over time \citep{varga2012scleroderma}. To systematically monitor skin
disease activity, clinicians use the TSS which is a quantitative score
between $0$ and $55$ computed by evaluating skin thickness at $17$
sites across the body (higher scores indicate more active skin
disease).

\vspace*{-7pt}
\subsection{Experimental Setup}
\label{sec:experimental-setup}
\vspace*{-7pt}

For our experiments, we extract trajectories from one of nation's
largest scleroderma patient registries. For both PFVC and TSS, we
study the trajectory from the time of first symptom until ten years of
follow-up. The PFVC dataset contains trajectories for 2,323
individuals and the TSS dataset contains 2,239 individuals. The median
number of observations per individuals is 3 for the PFVC data and 2
for the TSS data. The maximum number of observations is 55 and 22 for
PFVC and TSS respectively.

We present two sets of results. In the first, we visualize groups of
similar trajectories obtained by clustering the representations
learned by DTM. Although not quantitative, we use these visualizations
as a way to check that the DTM uncovers subpopulations that are
consistent with what is currently known about scleroderma. In the
second set of results, we use the learned representations of
trajectories obtained using the LMM, the reduced-rank model (FPCA) as
described by \citet{james2000principal}, and the DTM to statistically
test for relationships between important clinical outcomes and learned
disease trajectory representations.

For all experiments and all models, we use a common 5-dimensional
B-spline basis composed of degree-2 polynomials (see e.g. Chapter 20
in \citet{gelman2014bayesian}). We choose knots using the percentiles
of observation times across the entire training set
\citep{ramsay2002applied}. For the LMM and FPCA models, we use EM to
fit model parameters. To fit the DTM, we use the LMM estimate to set
the mean $\mu$ , noise $\sigma^2$, and average the diagonal elements
of $\matr{\Sigma}$ to set the kernel scale $\alpha$. Length-scales
$\ell$ are set to 1. For these experiments, we do not learn the
hyperparameters during optimization. We initialize the variational
means over $\vect{x}_i$ using the first two unit-scaled principal
components of $\vect{w}_i$ and set the variational covariances to be
diagonal with standard deviation $0.1$. For both PFVC and TSS, we use
minibatches of size $25$ and learn for a total of five epochs (passes
over the training data). The initial learning rate for $\vect{m}$ and
$\matr{S}$ is $0.1$ and decays as $t^{-1}$ for each epoch $t$.

\vspace*{-9pt}
\subsection{Qualitative Analysis of Representations}
\label{sec:qualitative-analysis-of-representations}

\vspace*{-5pt}
\begin{figure}
  \centering
  \includegraphics[width=1.0\linewidth]{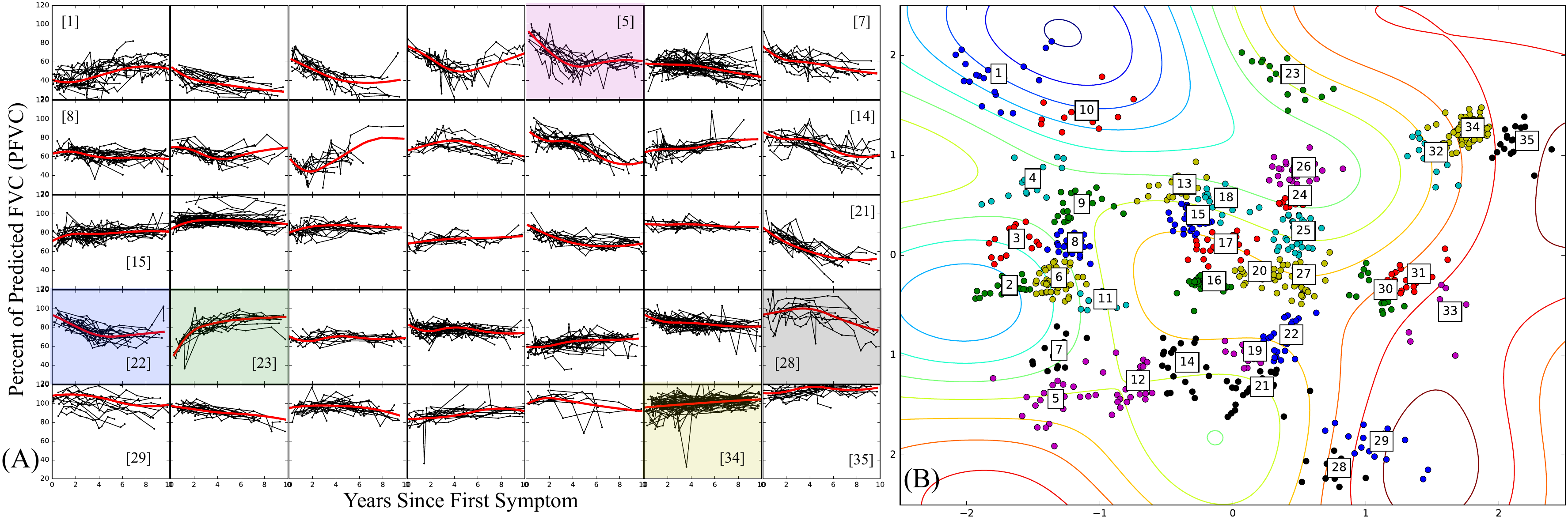}
  \vspace*{-20pt}
  \caption{\small{(A) Groups of PFVC trajectories obtained by hierarchical
    clustering of DTM representations. (B) Trajectory representations
    are color-coded and labeled according to groups shown in
    (A). Contours reflect posterior GP over the second B-spline
    coefficient (blue contours denote smaller values, red denote
    larger values).}}
  \label{fig:pfvc-map}
\end{figure}

\begin{figure}
  \centering
  \includegraphics[width=1.0\linewidth]{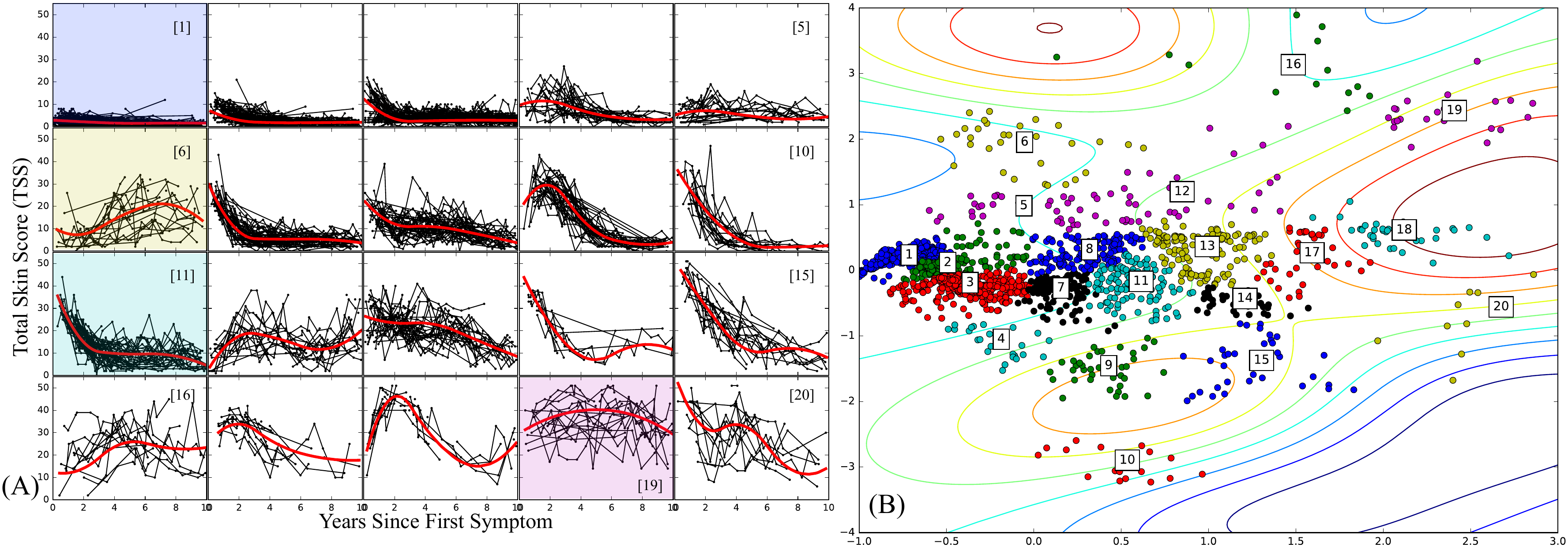}
  \vspace*{-20pt}
  \caption{\small{Same presentation as in Figure \ref{fig:pfvc-map} but for TSS trajectories.}}
  \label{fig:tss-map}
\end{figure}

\vspace*{-4pt} 
The DTM returns approximate posteriors over the
representations $\vect{x}_i$ for all individuals in the training
set. We examine these posteriors for both the PFVC and TSS datasets to
check for consistency with what is currently known about scleroderma
disease trajectories. In Figure~\ref{fig:pfvc-map}~(A) we show groups
of trajectories uncovered by clustering the learned representations,
which are plotted in Figure~\ref{fig:pfvc-map}~(B). Many of the groups
shown here align with other work on scleroderma lung disease subtypes
(e.g. \citet{schulam2015clustering}). In particular, we see rapidly
declining trajectories (group [5]), slowly declining trajectories
(group [22]), recovering trajectories (group [23]), and stable
trajectories (group [34]). Surprisingly, we also see a group of
individuals who we describe as ``late decliners'' (group [28]). These
individuals are stable for the first 5-6 years, but begin to decline
thereafter. This is surprising because the onset of
scleroderma-related lung disease is currently thought to occur early
in the disease course \citep{varga2012scleroderma}. In
Figure~\ref{fig:tss-map}~(A) we show clusters of TSS trajectories and
the corresponding color-coded representations in
Figure~\ref{fig:tss-map}~(B). These trajectories corroborate what is
currently known about skin disease in scleroderma. In particular, we
see individuals who have minimal activity (e.g. group [1]) and
individuals with early activity that later stabilizes (e.g. group
[11]), which correspond to what is known as the limited and diffuse
variants of scleroderma~\citep{varga2012scleroderma}. We also find
that there are a number of individuals with increasing activity over
time (group [6]) and some whose activity remains high over the ten
year period (group [19]). These patterns are not currently considered
to be canonical trajectories and warrant further investigation.

\vspace*{-9pt}
\subsection{Associations between Representations and Clinical Outcomes}
\label{sec:assocations-between-representations-and-clinical-outcomes}

\begin{table}[t]
  \scriptsize
  \caption{Disease Trajectory Held-out Log-Likelihoods}
  \label{tab:log-likelihoods}
  \centering
  \begin{tabular}{l rr rr}
    \toprule
          & \multicolumn{2}{c}{PFVC}                           & \multicolumn{2}{c}{TSS}                            \\
    \midrule
    Model & Subj. LL                 & Obs. LL                 & Subj. LL                 & Obs. LL                 \\
    \midrule
    LMM   & \tabresult{-17.59}{1.18} & \tabresult{-3.95}{0.04} & \tabresult{-13.63}{1.41} & \tabresult{-3.47}{0.05} \\
    FPCA  & \tabresult{-17.89}{1.19} & \tabresult{-4.03}{0.02} & \tabresult{-13.76}{1.42} & \tabresult{-3.47}{0.05} \\
    DTM   & \tabresult{-17.74}{1.23} & \tabresult{-3.98}{0.03} & \tabresult{-13.25}{1.38} & \tabresult{-3.32}{0.06} \\
    \bottomrule 
  \end{tabular}
\end{table}

\vspace*{-10pt}
\begin{table}[t]
  \scriptsize
  \caption{\small{P-values under the null hypothesis that 
    the distributions of trajectory representations are
    the same across individuals with and without clinical
    outcomes. Lower values indicate stronger support
    for rejection.}}
    \vspace{-6pt}
  \label{tab:outcome-associations}
  \centering
  \begin{tabular}{l rrr rrr}
    \toprule
                                    & \multicolumn{3}{c}{PFVC}                 & \multicolumn{3}{c}{TSS}                  \\
    \midrule
    Outcome                         &         LMM &         FPCA &         DTM &         LMM &         FPCA &         DTM \\
    \midrule
    Congestive Heart Failure        &       0.170 &        0.081 &       0.013 &       0.107 &        0.383 &       0.189 \\
    Pulmonary Hypertension          &       0.270 &    $^*$0.000 &   $^*$0.000 &       0.485 &        0.606 &       0.564 \\
    Pulmonary Arterial Hypertension &       0.013 &        0.020 &   $^*$0.002 &       0.712 &        0.808 &       0.778 \\
    Gastrointestinal Complications  &       0.328 &        0.073 &       0.347 &       0.026 &        0.035 &       0.011 \\
    Myositis                        &       0.337 &    $^*$0.002 &   $^*$0.004 &   $^*$0.000 &    $^*$0.002 &   $^*$0.000  \\
    Interstitial Lung Disease       &   $^*$0.000 &    $^*$0.000 &   $^*$0.000 &       0.553 &        0.515 &       0.495  \\
    Ulcers and Gangrene             &       0.410 &        0.714 &       0.514 &       0.573 &        0.316 &   $^*$0.009  \\
    \bottomrule
  \end{tabular}
\end{table}

\begin{figure}
  \centering
  \includegraphics[width=1.0\linewidth]{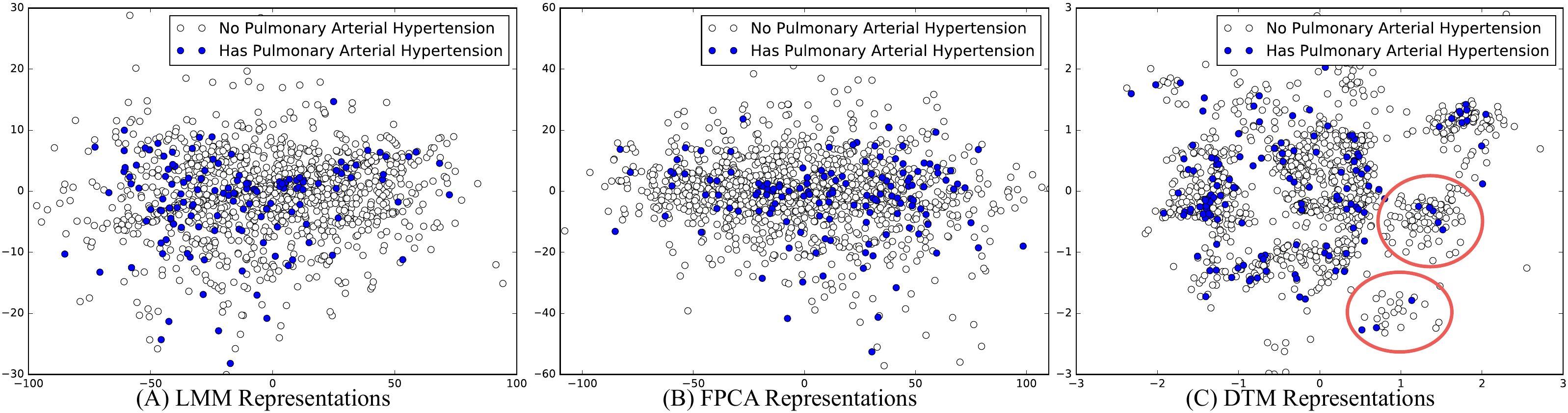}
  \caption{\small{Scatter plots of representations for the three models
    color-coded by presence or absence of pulmonary arterial
    hypertension (PAH). Groups of trajectories with very few cases of
    PAH are circled in red.}}
  \label{fig:outcomes-scatter}
\end{figure}

\vspace*{1pt}
To quantitatively evaluate the low-dimensional representations learned
by the DTM, we statistically test for relationships between the
representations of clinical marker trajectories and important clinical
outcomes. We compare the inferences of the hypothesis test with those
made using representations derived from the LMM and FPCA
baselines. For the LMM, we project $\vect{w}_i$ into its 2-dimensional
principal subspace. For FPCA, we learn a rank-2 covariance, which
recovers 2-dimensional embeddings. To establish that the models are
all equally expressive and achieve comparable generalization error, we
present held-out data log-likelihoods in
Table~\ref{tab:log-likelihoods}, which are estimated using 10-fold
cross-validation. We see that the models are roughly equivalent with
respect to generalization error.

To test associations between clinical outcomes and learned
representations, we use a kernel density estimator test
\citep{duong2012closed} to test the null hypothesis that the
distributions across subgroups with and without the outcome are
equivalent. The $p$-values obtained are listed in
Table~\ref{tab:outcome-associations}. As a point of reference, we
include two clinical outcomes that should be clearly related to the
two clinical markers. Interstitial lung disease is the most common
cause of lung damage in scleroderma \citep{varga2012scleroderma}, and
so we confirm that the null hypothesis is rejected for all three
models. Similarly, for TSS we expect ulcers and gangrene to be
associated with severe skin disease. In this case, only the
representations learned by DTM reveal this relationship. For the
remaining outcomes, we see that FPCA and DTM reveal similar
associations, but that only DTM suggests a relationship with pulmonary
arterial hypertension (PAH). Presence of fibrosis (which drives lung
disease progression) has been shown to be a risk factor in the
development of PAH (see Chapter 36 of \citet{varga2012scleroderma}), but
only the representations learned by DTM corroborate this finding (see
Figure~\ref{fig:outcomes-scatter}).

\vspace*{-5pt}
\vspace*{-5pt}
\section{Conclusion}
\label{sec:conclusion}
\vspace*{-9pt}

We present the Disease Trajectory Map (DTM), a novel probabilistic
model that learns low-dimensional embeddings of sparse and irregularly
sampled clinical time series data. The DTM is a reformulation of the
LMM that places an emphasis on the \emph{representations} that the
model learns. This view is comparable to that taken by
\citet{lawrence2004gaussian} in deriving the Gaussian process latent
variable model (GPLVM) from probabilistic principal component analysis
(PPCA)~\citep{tipping1999probabilistic}, and indeed the DTM can be
interpreted as a ``twin kernel'' GPLVM (briefly discussed in the
concluding paragraphs) over functional observations. The DTM can also
be viewed as an LMM with a ``warped'' Gaussian prior over the random
effects (see e.g. \citet{damianou2015variational} for a discussion of
distributions induced by mapping Gaussian random variables through
non-linear maps). We demonstrate the model by analyzing data extracted
from one of the nation's largest scleroderma patient registries, and
found that the DTM induces structure among trajectories that is
consistent with previous findings and also uncovers several surprising
disease trajectory shapes. We also explore associations between
important clinical outcomes and the DTM's representations and found
statistically significant differences in representations between
outcome-defined groups that were not uncovered by two sets of baseline
representations.

\newpage
\begingroup
\begin{small}
\bibliographystyle{plainnat}
\bibliography{main}
\end{small}
\endgroup

\appendix

\section{Derivation of Evidence Lower Bound}
\label{sec:derivation-of-evidence-lower-bound}

When marginalizing over the rows of $\matr{F}$, we induced a Gaussian
process over the trajectories, but by doing so we implicitly induced a
Gaussian process over the subject-specific basis coefficients. Let
$\vect{w}_i \triangleq \matr{F} \vect{x}_i \in \Reals^d$ denote the
curve weights implied by the mapping $\matr{F}$ and representation
$\vect{x}_i$, and let $\vect{w}_{:,k}$ for $k \in [d]$ denote the
$k^\text{th}$ coefficient of all subjects in the dataset. After
marginalizing the $k^\text{th}$ row of $\matr{F}$ and applying the
kernel trick, we see that the vector of coefficients $\vect{w}_{:,k}$
has a Gaussian process distribution with mean $0$ and covariance
\begin{align}
  \Cov(w_{ik}, w_{jk}) = \alpha k(\vect{x}_i, \vect{x}_j).
\end{align}
Moreover, the Gaussian processes across coefficients are mutually
statistically independent of one another. To construct our approximate
objective, we first approximate each of the $d$ coefficient Gaussian
processes by introducing $p$ inducing points (see
e.g. \cite{snelson2005sparse,titsias2009variational}) with values
$\vect{u}_k \in \Reals^p$ for each $k \in [d]$ observed at common
inputs $\vect{z}_i \in \Reals^q$ for $i \in [p]$. We assume that each
$\vect{w}_{:,k}$ and $\vect{u}_k$ are sampled from a common Gaussian
process, which implies the joint distribution:
\begin{align}
  \vect{u}_k &\given \vect{\theta} \sampled \Normal(\vect{0}, \matr{K}_{pp}) \\
  \vect{w}_k &\given \vect{u}_k, \vect{\theta}
  \sampled \Normal(\matr{K}_{mp} \inv{\matr{K}}_{pp} \vect{u}_k, \tilde{\matr{K}}_{mm}).
\end{align}
where $\matr{K}_{pp}$ is the Gram matrix between inducing points,
$\matr{K}_{mm}$ is the Gram matrix between subjects (based on their
representations $\vect{x}_i$), $\matr{K}_{mp}$ is the cross Gram
matrix between subjects and inducing points, and
$\tilde{\matr{K}}_{mm} \triangleq \matr{K}_{mm} - \matr{K}_{mp}
\inv{\matr{K}}_{pp} \matr{K}_{pm}$.

Now, we stack the inducing point values $\vect{u}_{1:d}$ into the
columns of a matrix
$\matr{U} \triangleq [ \vect{u}_1, \ldots, \vect{u}_d ]$. We will use
$\vect{u}$ to denote the ``vectorization'' of $\matr{U}$ obtained by
stacking the columns. Each row $i$ of $\matr{U}$ can be thought of as
the vector of coefficients belonging to a single \emph{inducing
  subject} which has an associated representation
$\vect{z}_i \in \Reals^q$. Let
$\vect{y} \triangleq \tra{[\tra{\vect{y}}_1, \ldots,
  \tra{\vect{y}}_m]}$ be the vector of concatenated trajectories and
$\matr{W}$ be the matrix containing subject $i$'s coefficients
$\vect{w}_i$ in each row, then following the derivation of
\citet{hensman2013gaussian}, we can lower bound the conditional
log-probability of $\vect{y}$ given $\vect{u}$ and $\vect{x}_{1:m}$:
\begin{align}
  \log p(\vect{y} \given \vect{u}, \vect{x}_{1:m})
  &= \log \int p(\vect{y} \given \matr{W}) p(\matr{W} \given \vect{u}, \vect{x}_{1:m}) d \matr{W} \\
  &= \log \int \prod_{i=1}^m p(\vect{y}_i \given \vect{w}_i) p(\matr{W} \given \vect{u}, \vect{x}_{1:m}) d \matr{W} \\
  &\ge \int p(\matr{W} \given \vect{u}, \vect{x}_{1:m}) \sum_{i=1}^m \log p(\vect{y}_i \given \vect{w}_i) d \matr{W} \\
  &= \sum_{i=1}^m \Exp_{p(\vect{w}_i \given \vect{u}, \vect{x}_i)}[ \log p(\vect{y}_i \given \vect{w}_i) ].
\end{align}
The expectation in each summand is easy to calculate because the mean
of $\vect{y}_i$ is linearly dependent on $\vect{w}_i$ and because the
conditional distribution $\vect{w}_i$ given $\vect{u}$ is multivariate
normal. Specifically, we have that
\begin{align}
  \vect{w}_i \given \vect{u}, \vect{x}_i \sampled
  \Normal( \tmatr{U} \imatr{K}_{pp} \vect{k}_i, \tilde{k}_{ii} \eye{d} ),
\end{align}
where $\vect{k}_i$ is a column vector filled with the $i^{\text{th}}$
row of $\matr{K}_{mp}$ and $\tilde{k}_{ii}$ is the $i^{\text{th}}$
diagonal element of $\tilde{\matr{K}}_{mm}$. Together with the
conditional distribution of $\vect{y}_i$ given $\vect{w}_i$, we have
that each summand can be written as
\begin{align}
  &\Exp_{p(\vect{w}_i \given \vect{u}, \vect{x}_i)}[ \log p(\vect{y}_i \given \vect{w}_i) ] \\
  &= -\frac{n_i}{2} \log 2\pi \sigma^2
  - \frac{1}{2\sigma^2} \Exp_{p(\vect{w}_i \given \vect{u}, \vect{x}_i)}[ \tra{(\vect{y}_i - \mu - \matr{B}_i \vect{w}_i)} (\vect{y}_i - \mu - \matr{B}_i \vect{w}_i) ] \\
  &= \log \Normal(\vect{y}_i \given \mu + \matr{B}_i \tmatr{U} \imatr{K}_{pp} \vect{k}_i, \sigma^2 \eye{n_i})
  - \frac{\tilde{k}_{ii}}{2\sigma^2} \Tr{\tmatr{B}_i \matr{B}_i} \\
  &\triangleq \log \tilde{p} (\vect{y}_i \given \vect{u}, \vect{x}_i).
\end{align}
We can now write the lower bound on the conditional log-probability as
\begin{align}
  \log p(\vect{y} \given \vect{u}, \vect{x}_{1:m})
  \ge \sum_{i=1}^m \log \tilde{p}(\vect{y}_i \given \vect{u}, \vect{x}_i)
  \triangleq \log \tilde{p}(\vect{y} \given \vect{u}, \vect{x}_{1:m}).
\end{align}

To complete the derivation of the approximate objective, we use the
lower bound on $\log p(\vect{y} \given \vect{u}, \vect{x}_{1:m})$ to
create a variational lower bound on the marginal log-probability of
the trajectories
\begin{align}
  \log p(\vect{y})
  &= \log \int p(\vect{y} \given \vect{u}, \vect{x}_{1:m}) p(\vect{u}, \vect{x}_{1:m}) d \vect{u} \\
  &\ge \int q(\vect{u}, \vect{x}_{1:m}) \left( \log p( \vect{y} \given \vect{u}, \vect{x}_{1:m} ) - \log q(\vect{u}, \vect{x}_{1:m}) + \log p(\vect{u}, \vect{x}_{1:m}) \right) d\vect{u} d\vect{x}_{1:m} \\
  &\ge \int q(\vect{u}, \vect{x}_{1:m}) \left( \log \tilde{p}( \vect{y} \given \vect{u}, \vect{x}_{1:m} ) - \log q(\vect{u}, \vect{x}_{1:m}) + \log p(\vect{u}, \vect{x}_{1:m}) \right) d\vect{u} d\vect{x}_{1:m} \\
  &\triangleq \log \tilde{p} (\vect{y}).
\end{align}
We assume that $\vect{u}$, $\vect{x}_1, \ldots, \vect{x}_m$ are all
mutually independent in the variational posterior. We use a
multivariate normal variational approximation for each $\vect{x}_i$
with variational parameters $\vect{m}_i$ and $\matr{S}_i$. 

Fixing $\vect{x}_i$, to find the the optimal form for $q(\vect{u})$,
note that each
$\log \tilde{p}(\vect{y}_i \given \vect{u}, \vect{x}_i)$ is composed
of a log-likelihood plus an additive term that is independent of
$\vect{u}$. Therefore, the terms that depend on $\vect{u}$ can be
written as:
\begin{align}
  \Exp_{q(\vect{u})} \left[ \sum_{i=1}^m \log \Normal(\vect{y}_i \given \mu + \matr{B}_i \tmatr{U} \imatr{K}_{pp} \vect{k}_i, \sigma^2 \eye{n_i}) \right] - \KL(q \| p).
\end{align}
Now, note that the mean in any of the log-likelihood terms can be
rewritten as
\begin{align}
  \mu + \matr{B}_i \tmatr{U} \imatr{K}_{pp} \vect{k}_i
  = \mu + \kronprod{\matr{B}_i}{\tvect{k}_i \imatr{K}_{pp}} \vect{u},
\end{align}
Let
$\matr{C}_i \triangleq \kronprod{\matr{B}_i}{\tvect{k}_i
  \imatr{K}_{pp}}$ denote the \emph{extended design matrix} obtained
through this rewriting, and recall that each column $\vect{u}_k$ is
normally distributed with mean zero and covariance
$\matr{K}_{pp}$. The prior over the vectorized matrix $\vect{u}$ is
therefore also multivariate normal.  The expression above is maximized
when $q(\vect{u})$ is equal to the posterior over $\vect{u}$ given the
observed trajectories. Because the prior is multivariate normal and
the mean of the likelihood depends linearly on $\vect{u}$, the
posterior must also be multivariate normal. Moreover, we know its
exact form:
\begin{align}
  \vect{m}_* = \matr{S}_* \left( \sigma^{-2} \sum_{i=1}^m \tmatr{C}_i (\vect{y}_i - \mu) \right)
  \,\,\, , \,\,\,
  \matr{S}_* = \inv{\left( \sigma^{-2} \sum_{i=1}^m \tmatr{C}_i \matr{C}_i + \kronprod{\eye{d}}{\imatr{K}_{pp}} \right)}.
\end{align}
We therefore parameterize $q(\vect{u})$ as a multivariate normal
distribution with variational parameters $\vect{m}$ and $\matr{S}$.

We now derive a closed-form expression for the expectation of
$\log \tilde{p}(\vect{y}_i \given \vect{u}, \vect{x}_i)$ under
variational posterior distribution. Because $\vect{u}$ and
$\vect{x}_i$ are assumed to be independent in the variational
posteriors, we can analyze the expectation in either order. Fix
$\vect{x}_i$, then we see that
$\log \tilde{p}(\vect{y}_i \given \vect{u}, \vect{x}_i)$ depends on
$\vect{u}$ only through the mean of the Gaussian density, which is a
quadratic term in log likelihood. Because $q(\vect{u})$ is
multivariate normal, we can compute the expectation in closed form.
\begin{align*}
  \Exp_{q(\vect{u})} [ \log \tilde{p}(\vect{y}_i \given \vect{u}, \vect{x}_i) ]
  &= \Exp_{q(\matr{U})} [ \log \Normal(\vect{y}_i \given \mu + \kronprod{\matr{B}_i}{\tvect{k}_i \imatr{K}_{pp}} \vect{u}, \sigma^2 \eye{n_i}) ]
     - \frac{\tilde{k}_{ii}}{2 \sigma^2} \Tr{\tmatr{B}_i \matr{B}_i} \\
  &= \log \Normal(\vect{y}_i \given \mu + \matr{C}_i \vect{m}, \sigma^2 \eye{n_i}) ]
     - \frac{1}{2 \sigma^2} \Tr{\matr{S} \tmatr{C}_i \matr{C}_i}
     - \frac{\tilde{k}_{ii}}{2 \sigma^2} \Tr{\tmatr{B}_i \matr{B}_i},
\end{align*}
We can compute the expectation of
$\Exp_{q(\vect{u})}[\log \tilde{p}(\vect{y}_i \given \vect{u},
\vect{x}_i)]$ in closed form by noting that we need only compute
expectations of $\vect{k}_i$ and $\vect{k}_i
\tvect{k}_i$. Specifically, we have that
\begin{align}
  \Exp_{q(\vect{x}_i)}[ k(\vect{x}_i, \vect{z}_j) ]
  = \frac{\alpha}{|\matr{S}_i|^{1/2} |\matr{A}|^{1/2}} \exp \left\{ \frac{1}{2} (\tvect{B} \imatr{A} \vect{b} - c) \right\},
\end{align}
where $\matr{A} = \imatr{S}_i + \ell^{-2} \eye{q}$,
$\vect{b} = \imatr{S}_i \vect{m}_i + \ell^{-2} \vect{z}_j$, and
$c = \tvect{m}_i \imatr{S}_i \vect{m} + \ell^{-2} \tvect{z}_j
\vect{z}_j$. Similarly, for the expected outer product, we have
\begin{align}
  \Exp_{q(\vect{x}_i)}[ k(\vect{x}_i, \vect{z}_j) k(\vect{x}_i, \vect{z}_k) ]
  = \frac{\alpha}{|\matr{S}_i|^{1/2} |\matr{A}|^{1/2}} \exp \left\{ \frac{1}{2} (\tvect{B} \imatr{A} \vect{b} - c) \right\},
\end{align}
where $\matr{A} = \imatr{S}_i + 2 \ell^{-2} \eye{q}$,
$\vect{b} = \imatr{S}_i \vect{m}_i + \ell^{-2} \vect{z}_j + \ell^{-2}
\vect{z}_k$, and
$c = \tvect{m}_i \imatr{S}_i \vect{m} + \ell^{-2} \tvect{z}_j
\vect{z}_j + \ell^{-2} \tvect{z}_k \vect{z}_k$. Importantly, we can
simply substitute these expectations into
$\Exp_{q(\vect{u})}[\log \tilde{p}( \vect{y}_i \given \vect{u},
\vect{x}_i )]$ and the form of the lower bound does not change (it is
still a Gaussian log-likelihood plus the additional trace terms).

\section{Optimizing the Evidence Lower Bound}
\label{sec:optimizing-the-approximate-objective}

To formulate the complete objective, we use the lower bound derived
above and place priors on the observation noise $\sigma^2$, and the
hyperparameters of the kernel $k(\cdot, \cdot)$. In this section and
in our experiments we assume that the kernel is a radial basis
function (RBF) with scale $\alpha$ and length-scale (or bandwidth)
$\ell$. We assumelog normal distributions over $\sigma^2$, $\alpha$,
and $\ell$ with mean parameters $m_s$, $m_a$, $m_\ell$ respectively
and precision parameters $\rho_s$, $\rho_a$, and $\rho_\ell$
respectively. Our objective is therefore
\begin{align}
  \mathcal{J}_{\text{SA-DTM}}( & \vect{m}, \matr{S}, \vect{m}_{1:m}, \matr{S}_{1:m}, \mu, \sigma^2, \alpha, \ell) = \\
  &\phantom{{}+{}} \sum_{i=1}^m -\frac{n_i}{2} \log 2 \pi \sigma^2 - \frac{1}{2\sigma^2} \Exp_{q(\vect{x}_i)}[ \| \vect{y}_i - \mu - \kronprod{\matr{B}_i}{\tvect{k}_i \imatr{K}_{pp}} \vect{m} \|_2^2 ] \\
  &+ \sum_{i=1}^m -\frac{1}{2\sigma^2} \Tr{ \matr{S} \kronprod{ \tmatr{B}_i \matr{B}_i }{ \imatr{K}_{pp} \Exp_{q(\vect{x}_i)}[\vect{k}_i \tvect{k}_i] \imatr{K}_{pp} } } \\
  &+ \sum_{i=1}^m -\frac{1}{2\sigma^2} \Tr{ \tmatr{B}_i \matr{B}_i } ( \alpha - \Exp_{q(\vect{x}_i)}[\tvect{k}_i \imatr{K}_{pp} \vect{k}_i] ) \\
  &- \sum_{i=1}^m \frac{1}{2} \left( \Tr{\matr{S}_i + \vect{m}_i \tvect{m}_i} - q - \log |\matr{S}_i| \right) \\
  &- \frac{1}{2} \left( \Tr{ (\matr{S} + \vect{m} \tvect{m}) \kronprod{\eye{d}}{\imatr{K}_{pp}} } - p d + \log \frac{|\matr{K}_{pp}|^d}{|\matr{S}|} \right) \\
  &- \frac{\rho_s}{2} \| \log \sigma^2 - m_s \|^2_2 - \frac{\rho_a}{2} \| \log \alpha - m_a \|^2_2 - \frac{\rho_\ell}{2} \| \log \ell - m_\ell \|^2_2.
\end{align}
Note that the last three lines above can be seen as regularizers (log
priors for the hyperparameters and a $\KL$ divergence between the
variational distribution $q$ and the prior $p$). The first four lines
can be decomposed across subjects, suggesting that we can use
stochastic approximation of the objective and its gradients to derive
a scalable algorithm for optimizing the objective.

We define an iterative first-order optimization algorithm. In broad
strokes, within each iteration we will sample a single subject $i$ (or
a batch of patients), maximize the objective with respect to
$\vect{m}_i$ and $\matr{S}_i$ while holding the global variables
fixed, compute the approximate gradients of the objective, and take a
small step in the direction of each gradient for each parameter (the
step size is determined by a learning schedule, which may be specific
to each global variable). We discuss each step in detail below. We do
so assuming a single sampled subject $i$, although in principle we can
sample a batch of subjects to reduce variance in the gradient
estimate.

\paragraph{Maximizing wrt local variables ($\vect{m}_i, \matr{S}_i$).}
Before computing gradients of the approximate objective with respect
to the global parameters, we first do a block coordinate optimization
over the local variational parameters of subject $i$. We optimize:
\begin{align}
  J_i & (\vect{x}_i) = \\
  &- \frac{n_i}{2} \log 2 \pi \sigma^2 - \frac{1}{2 \sigma^2} \Exp_{q(\vect{x_i})} [\| \vect{y}_i - \mu - \kronprod{\matr{B}_i}{\tvect{k}_i \imatr{K}_{pp}} \vect{m} \|^2_2] \\
  &- \frac{1}{2 \sigma^2} \Tr{ \matr{S} \kronprod{ \tmatr{B}_i \matr{B}_i }{ \imatr{K}_{pp} \Exp_{q(\vect{x_i})} [\vect{k}_i \tvect{k}_i] \imatr{K}_{pp} } } \\
  &- \frac{1}{2 \sigma^2} \Tr{\tmatr{B}_i \matr{B}_i} (\alpha - \Exp_{q(\vect{x_i})} [\tvect{k}_i \imatr{K}_{pp} \vect{k}_i]).
\end{align}
We can optimize this expression using a gradient-based optimizer. We
use the scaled conjugate gradients algorithm.

\paragraph{Estimating gradients of global variables.}
Having sampled subject $i$ and refit her local variational parameters,
we now want to approximate the gradient of the full objective with
respect to the global variables $\vect{m}$, $\matr{S}$, $\mu$,
$\sigma^2$, $\alpha$, and $\ell$. We first look at the approximate
gradient with respect to $\vect{m}$.
\begin{align}
  \hat{\nabla}_{\mathcal{J}_{\text{SA-DTM}}}(\vect{m}) = 
  \Exp_{q(\vect{x}_i)}[\frac{m}{\sigma^2} \kronprod{\tmatr{B}_i}{\imatr{K}_{pp} \vect{k}_i} (\vect{y}_i - \mu - \kronprod{\matr{B}}{\tvect{k}_i \imatr{K}_{pp}} \vect{m})]
  - \kronprod{\eye{d}}{\imatr{K}_{pp}} \vect{m}.
\end{align}
The approximate gradient with respect to $\matr{S}$ is
\begin{align}
  \hat{\nabla}_{\mathcal{J}_{\text{SA-DTM}}}(\matr{S}) = 
  &- \frac{m}{2\sigma^2} \Tr{ \kronprod{\tmatr{B}_i \matr{B}_i}{\imatr{K}_{pp} \Exp_{q(\vect{x}_i)}[\vect{k}_i \tvect{k}_i] \imatr{K}_{pp}} } \\
  &- \frac{1}{2} \Tr{ \kronprod{\eye{d}}{\imatr{K}_{pp}} }
   + \frac{1}{2} \Tr{ \imatr{S} }.  
\end{align}

Note that if we set these approximate gradients to $0$, we obtain the
following estimates of $\vect{m}$ and $\matr{S}$:
\begin{align}
  \hat{\vect{m}} &= \hat{\matr{S}} \left( \frac{m}{\sigma^2} \kronprod{\tmatr{B}_i}{\imatr{K}_{pp} \Exp_{q(\vect{x}_i)}[\vect{k}_i]} (\vect{y} - \mu) \right) \\
  \hat{\matr{S}} &= \left( \frac{m}{\sigma^2} \kronprod{\tmatr{B}_i \matr{B}_i}{\imatr{K}_{pp} \Exp_{q(\vect{x}_i)}[\vect{k}_i \tvect{k}_i] \imatr{K}_{pp}} + \kronprod{\eye{d}}{\imatr{K}_{pp}} \right)^{-1}
\end{align}
We can improve the rate of convergence of our algorithm by taking the
geometry of the space of distributions parameterized by $\vect{m}$ and
$\matr{S}$ into account. We do so by using the \emph{natural
  gradients} for these two parameters instead of the approximations
above. Let $\vect{\theta}_1$ and $\matr{\theta}_2$ denote the
canonical parameterization of the variational multivariate normal,
then the gradient updates at time $t$ are \cite{hoffman2013stochastic}:
\begin{align}
  \vect{\theta}_1^t &= \vect{\theta}_1^{t-1} + \lambda_t ( \eta_1^{t-1} - \vect{\theta}_1^{t-1} ) \\
  \matr{\theta}_2^t &= \matr{\theta}_2^{t-1} + \lambda_t ( \eta_2^{t-1} - \matr{\theta}_2^{t-1} ),
\end{align}
where
\begin{align}
  \eta_1^{t-1} &= \frac{m}{\sigma^2} \kronprod{\tmatr{B}_i}{\imatr{K}_{pp} \Exp_{q(\vect{x}_i)}[\vect{k}_i]} (\vect{y} - \mu) \\
  \eta_2^{t-1} &= -\frac{m}{2 \sigma^2} \kronprod{\tmatr{B}_i \matr{B}_i}{\imatr{K}_{pp} \Exp_{q(\vect{x}_i)}[\vect{k}_i \tvect{k}_i] \imatr{K}_{pp}}
\end{align}

To update the hyperparamters, we need to compute the gradients with
respect to $\mu$, $\sigma^2$, $\alpha$, and $\ell$. We parameterize
$\sigma^2$, $\alpha$, and $\ell$ using their logarithms, and so
present gradients with respect to that representation. To make the
expressions more clear, we present the gradients as differentials with
respect to the kernel, which can be completed using the chain
rule. The estimate of the gradient with respect to $\mu$ is
\begin{align}
  \hat{\nabla}_{\mathcal{J}_{\text{SA-DTM}}}(\mu) = 
  \frac{m}{\sigma^2} (\vect{y}_i - \mu - \kronprod{\matr{B}_i}{\expkt{} \imatr{K}_{pp}}\vect{m})^\T \onesv{n_i}.
\end{align}
The estimate of the gradient with respect to $\log \sigma^2$ is
\begin{align}
  \hat{\nabla}_{\mathcal{J}_{\text{SA-DTM}}}(\log \sigma^2) =
  &- \frac{m n_i}{2} + \frac{m}{2\sigma^2} \Exp_{q(\vect{x}_i)}[\| \vect{y}_i - \mu - \kronprod{\matr{B}_i}{\tvect{k}_i \imatr{K}_{pp}}\vect{m} \|^2_2] \\
  &+ \frac{m}{2\sigma^2} \Tr{ \matr{S} \kronprod{\tmatr{B}_i \matr{B}_i}{\imatr{K}_{pp} \expkkt{} \imatr{K}_{pp}} } \\
  &+ \frac{m}{2\sigma^2} \Tr{\tmatr{B}_i \matr{B}} (\alpha - \Tr{ \imatr{K}_{pp} \expkkt{} }) \\
  &- \rho_s (\log \sigma^2 - m_s).  
\end{align}
The estimate of the gradient with respect to $\log \alpha$ is
\begin{align}
  \hat{\nabla}_{\text{SA-DTM}}& ( \log \alpha) = \\
  &\phantom{{}+{}} \frac{m}{\sigma^2} \Exp_{q(\vect{x}_i)}[(\vect{y}_i - \mu - \matr{C} \vect{m})^\T \kronprod{\matr{B}_i}{\partial \tvect{k}_i \imatr{K}_{pp} - \tvect{k}_i \imatr{K}_{pp} \partial \matr{K}_{pp} \imatr{K}_{pp}} \vect{m}] \\
  &- \frac{m}{\sigma^2} \Exp_{q(\vect{x}_i)}[\Tr{\matr{S} \tmatr{C}_i \kronprod{\matr{B}_i}{\partial \tvect{k}_i \imatr{K}_{pp} - \tvect{k}_i \imatr{K}_{pp} \partial \matr{K}_{pp} \imatr{K}_{pp}}}] \\
  &- \frac{m}{2 \sigma^2} \Tr{\tmatr{B}_i \matr{B}_i} \alpha \\
  &+ \frac{m}{2 \sigma^2} \Tr{\tmatr{B}_i \matr{B}_i} (2 \expkt{} \imatr{K}_{pp} \partial \expk{} - \Tr{ \imatr{K}_{pp} \partial \matr{K}_{pp} \imatr{K}_{pp} \expkkt{} }) \\
  &+ \frac{1}{2} \left( \Tr{(\matr{S} + \matr{m} \tmatr{m}) \kronprod{\eye{d}}{\imatr{K}_{pp} \partial \matr{K}_{pp} \imatr{K}_{pp}}} - d \Tr{\imatr{K}_{pp} \partial \matr{K}_{pp}} \right).
\end{align}
The estimate of the gradient with respect to $\log \ell$ is
\begin{align}
  \hat{\nabla}_{\text{SA-DTM}}& ( \log \ell) = \\
  &\phantom{{}+{}} \frac{m}{\sigma^2} \Exp_{q(\vect{x}_i)}[(\vect{y}_i - \mu - \matr{C} \vect{m})^\T \kronprod{\matr{B}_i}{\partial \tvect{k}_i \imatr{K}_{pp} - \tvect{k}_i \imatr{K}_{pp} \partial \matr{K}_{pp} \imatr{K}_{pp}} \vect{m}] \\
  &- \frac{m}{\sigma^2} \Exp_{q(\vect{x}_i)}[\Tr{\matr{S} \tmatr{C}_i \kronprod{\matr{B}_i}{\partial \tvect{k}_i \imatr{K}_{pp} - \tvect{k}_i \imatr{K}_{pp} \partial \matr{K}_{pp} \imatr{K}_{pp}}}] \\
  &+ \frac{m}{2 \sigma^2} \Tr{\tmatr{B}_i \matr{B}_i} (2 \expkt{} \imatr{K}_{pp} \partial \expk{} - \Tr{ \imatr{K}_{pp} \partial \matr{K}_{pp} \imatr{K}_{pp} \expkkt{} }) \\
  &+ \frac{1}{2} \left( \Tr{(\matr{S} + \matr{m} \tmatr{m}) \kronprod{\eye{d}}{\imatr{K}_{pp} \partial \matr{K}_{pp} \imatr{K}_{pp}}} - d \Tr{\imatr{K}_{pp} \partial \matr{K}_{pp}} \right).
\end{align}

\end{document}